\documentclass[conference]{IEEEtran}
\IEEEoverridecommandlockouts

\usepackage[T1]{fontenc}
\usepackage{cite}
\usepackage{amsmath,amssymb,amsfonts}
\usepackage{booktabs}
\usepackage{graphicx}
\usepackage{array}
\usepackage{microtype}
\usepackage{textcomp}
\usepackage{url}
\usepackage[hidelinks]{hyperref}
\title{EmotiQ: A Quantum Residual Specialist Framework for Text Emotion Classification\\
\thanks{This work was supported by the Lemann Student/Faculty Collaborative Research Fund at Earlham College.}
}

\author{\IEEEauthorblockN{Carlton Kigonya}
\IEEEauthorblockA{\textit{Department of Computer Science} \\
\textit{Earlham College}\\
Richmond, IN, USA \\
cskigon24@earlham.edu}
\and
\IEEEauthorblockN{Tanaka Chirara}
\IEEEauthorblockA{\textit{Department of Computer Science} \\
\textit{Earlham College}\\
Richmond, IN, USA \\
tchira24@earlham.edu}
\and
\IEEEauthorblockN{Yunting Yin}
\IEEEauthorblockA{\textit{Department of Computer Science} \\
\textit{Eastern Michigan University}\\
Ypsilanti, MI, USA \\
yyin@emich.edu}
}

\begin{document}
\maketitle

\begin{abstract}
Emotion classification from text is challenging because many affective states are distinguished by subtle linguistic signals and semantically overlapping boundaries. We investigate whether compact quantum modules can serve as targeted residual specialists that assist traditional classifiers on difficult emotion decisions. We introduce EmotiQ, a residual specialist framework for text emotion classification using a frozen transformer classifier as the primary decision maker, identifies challenging emotion boundaries from validation behavior, and trains boundary specialists, including quantum encodings, data re-uploading specialists, prototype and fidelity specialists, and kernel-based specialists. Specialists are permitted to intervene only when they satisfy validation-based admission criteria. We evaluate the framework across six text emotion classification datasets and establish traditional model baselines. Experimental results show that the admitted EmotiQ policy improves mean macro-F1 metrics over the baseline, and achieves positive gains on five of six datasets. Among the quantum components, ZZ interaction feature maps are the most consistently effective, yielding a mean macro-F1 improvement of +0.0030 and positive gains on four datasets. These findings suggest that current quantum NLP methods can be deployed as residual specialists that provide localized improvements under strong classical supervision.
\end{abstract}

\begin{IEEEkeywords}
quantum natural language processing, emotion classification, quantum computing
\end{IEEEkeywords}

\section{Introduction}
Emotion classification aims to identify affective states such as joy, anger, fear, sadness, surprise, and disgust from data. Despite substantial advances in language modeling, fine-grained emotion recognition remains challenging for textual data because many affective categories are separated by ambiguous decision boundaries. Closely related emotions often share vocabulary and sentiment, making them difficult to distinguish even for strong pretrained language models. These characteristics make emotion classification a useful benchmark for studying representation learning methods that can capture complex nonlinear relationships in textual data. Quantum natural language processing explores whether quantum-inspired representations can improve language understanding tasks. In this study, we ask whether these methods can provide complementary information for difficult low-confidence decisions.

To investigate this idea, we propose EmotiQ, a quantum residual specialist framework for text emotion classification. EmotiQ uses a frozen DistilBERT classifier as the primary decision maker and uses validation behavior to identify emotion boundaries where the base model is frequently confused or uncertain. For each selected boundary, the framework trains a bank of compact specialists, including quantum encodings, data re-uploading circuits, prototype-based models, and kernel-based methods \cite{schuld2019,huang2021}. Specialists are allowed to intervene only after satisfying validation-based admission criteria that evaluate predictive gain.

We evaluate the framework on six emotion classification datasets spanning different label granularities and domains.  Experimental results show that compact quantum modules can provide modest but consistent gains when deployed selectively on emotion boundaries discovered in validation. More broadly, the results suggest that current quantum NLP methods can be used as residual components operating alongside classical models. Code is available at \href{https://github.com/AcaiLab/EmotiQ-text}{https://github.com/AcaiLab/EmotiQ-text}.

\section{Related Work}
\textbf{Emotion classification.}
SemEval-2025 Task 11 expanded text emotion detection to multilingual,
multi-label, and intensity-prediction settings~\cite{semeval2025_emotion}.
Systems developed for this task include transformer ensembles and
contrastive training of large language models~\cite{qimp2025_emotion,chinchunmei2025_emotion,kim2026interpretabledepressiondetectionsocial}. Other approaches combine RoBERTa with recurrent models or lightweight
adapters to capture contextual and sequential information~\cite{ler2026,ramha2025}.
These studies motivate strong classical baselines. Our focus is a different
question: whether a compact specialist can improve selected decisions made
by a fixed base model.

\textbf{Quantum language models.}
Compositional models connect linguistic structure with quantum
representations~\cite{coecke2010}, and libraries such as
\emph{lambeq} support quantum NLP pipelines~\cite{lambeq}.
Hybrid approaches instead pass classical embeddings through quantum
feature maps, variational circuits, or kernels~\cite{havlicek2019,mari2020}.
Data re-uploading repeatedly encodes the same input between trainable
circuit operations~\cite{perezsalinas2020}. Recent work also examines
compact quantum models for natural language inference~\cite{sun2025}.
EmotiQ uses these ideas in a selective correction policy. It evaluates
quantum-inspired classical maps separately from simulated trainable quantum
circuits and includes classical nonlinear controls.

\section{Datasets and Evaluation}
We use six datasets with 6-20 emotion labels shown in Table~\ref{tab:datasets}. AI-Roadmap contains Reddit-style text with six
Ekman emotions. BRIGHTER is the English portion of a multilingual emotion
benchmark, Bulpara-8 covers eight emotions, and Super Emotion is a large
aggregate dataset. For these multi-label sources, we keep only examples
with one active emotion label. Emotion20 has a finer-grained 20-class label
set, and Tanaos-8 contains synthetic examples for eight emotions.

\begin{table}[t]
\centering
\caption{Dataset splits before experimental sampling.}
\label{tab:datasets}
\small
\setlength{\tabcolsep}{3.5pt}
\begin{tabular}{@{}lrrrr@{}}
\toprule
Dataset & Train & Val. & Test & Labels \\
\midrule
AI-Roadmap & 2,781 & 309 & 773 & 6 \\
BRIGHTER & 1,140 & 92 & 2,186 & 6 \\
Bulpara-8 & 26,276 & 3,320 & 3,330 & 8 \\
Emotion20 & 57,308 & 6,368 & 15,919 & 20 \\
Super Emotion & 431,371 & 53,824 & 56,339 & 7 \\
Tanaos-8 & 8,196 & 911 & 2,277 & 8 \\
\bottomrule
\end{tabular}
\end{table}

We summarize held-out performance with macro-F1, and also records accuracy in experiments. Macro-F1 averages each class's F1 score equally, which prevents frequent
emotions from dominating the result. Every reported gain is relative to the
frozen DistilBERT base evaluated under the same experiment's protocol.
Means give equal weight to the six datasets.

The dataset sizes in Table~\ref{tab:datasets} describe the prepared splits. To bound computation, the classical suite samples up to 1,800 training and 700 test examples per
dataset. Encoding screening, ambiguity heads, and the final framework use
caps of 700 training, 300 validation, and 300 test examples. The supporting
pair specialist experiment uses caps of 1,800/700/700. We balance the samples across emotion classes, using only the examples available in each split. We use validation data to identify difficult emotion pairs, choose model settings, and decide which specialists can intervene. Test labels are used only to evaluate the final predictions. Because these experiments use different samples and decision rules, raw F1 scores across sections are not directly
comparable. We instead focus on within-experiment gains.

\section{EmotiQ Framework}
\begin{figure*}[t]
\centering
\includegraphics[width=\textwidth]{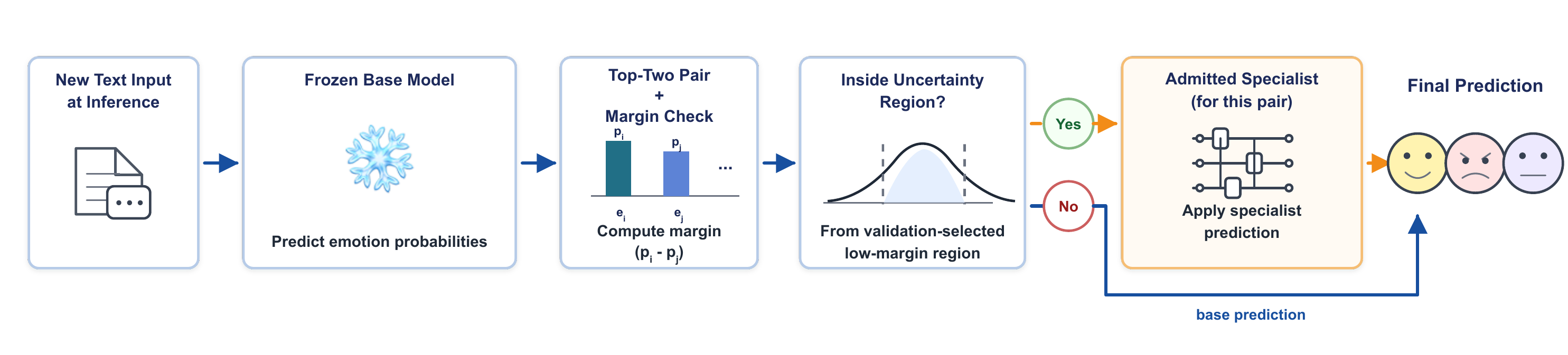}
\caption{EmotiQ at inference time. The base classifier produces emotion
probabilities. A matching, admitted specialist may intervene within its
validation-selected margin region; otherwise the base prediction is retained.}
\label{fig:emotiq-residual-path}
\end{figure*}

Figure~\ref{fig:emotiq-residual-path} shows the prediction path. The base
classifier handles every input first. A specialist can change its decision
only when the predicted emotion pair matches the specialist and the
confidence margin is at or below a threshold chosen on validation data.

\subsection{Base Model and Features}
We freeze the text encoders and train the classifier heads. DistilBERT
produces a 768-dimensional mean-pooled representation for the base logistic
regression model. The specialist branch uses a separate frozen MiniLM
encoder, producing 384-dimensional embeddings. For each emotion pair,
training embeddings are standardized, reduced to four dimensions by
principal component analysis (PCA), and normalized. The fitted transforms
are then applied to validation and test examples. Inputs are truncated to
96 tokens before embedding.

The final quantum specialists use four qubits, with one compressed value
encoded per qubit. Circuits are simulated with PennyLane's \texttt{default.qubit} backend and trained through PyTorch. Classical models and preprocessing use scikit-learn; text encoders are loaded through Hugging Face Transformers.

\subsection{Discovering Hard Emotion Boundaries}
A boundary is a pair of emotion labels that the base model struggles to
separate. On validation data, we score each candidate pair using four
signals: confusion in both directions, frequency among the top two
predictions, the proportion of small confidence margins, and a heuristic
for semantic similarity. The last signal, called \emph{semantic degeneracy}
in the implementation, groups related labels into broad affective families.

Let $p_{(1)}(x)$ and $p_{(2)}(x)$ be the two highest base probabilities. The
confidence margin is
\begin{equation}
 m(x)=p_{(1)}(x)-p_{(2)}(x).
\end{equation}
A small margin means that the base model has difficulty choosing between
the two labels. We retain up to three high-scoring pairs per dataset,
subject to training support. Examples include anger and disgust on AI-Roadmap,
joy and excitement on Tanaos-8, and joy and love on Super Emotion.

\subsection{Training Candidate Specialists}
Each pair receives a bank of small models trained on examples from its two
classes. An angle-encoded variational quantum circuit returns four
individual Pauli-$Z$ expectation values. A re-uploading circuit returns
these four values plus four neighboring $ZZ$ correlations. The quantum
outputs are concatenated with the four PCA features, giving 8-dimensional
inputs for the angle specialist's logistic head and 12-dimensional inputs
for the re-uploading head. Shared re-uploading variants reuse circuit
parameters across related boundaries while retaining pair-specific heads.

The bank also includes two lightweight classical alternatives. A prototype-fidelity model represents each class by its average feature vector, scaled to unit length. It then compares an input with each class representative, using squared inner products as similarity scores.
A SVM learns a nonlinear decision boundary in the compact feature space. These controls help distinguish a benefit from quantum circuits from a benefit due to local specialization more generally.

\subsection{Selecting Reliable Specialists}
For each candidate, validation selects a margin threshold. We then measure
the change in overall validation macro-F1 and examine the predictions that
actually changed. \emph{Override precision} is the fraction of changed
predictions that are correct after intervention. \emph{Harm rate} is the
fraction of changed predictions that were correct under the base model
but become wrong. A candidate is eligible only if it changes at least one
prediction and satisfies all four conditions:
\begin{align}
 \Delta F1_{\mathrm{val}} &> 0.0001, &
 \mathrm{precision} &\geq 0.65, \nonumber\\
 \mathrm{harm} &\leq 0.10, &
 \mathrm{gain/parameter} &\geq 0.
\end{align}
The last condition checks nonnegative utility using the implementation's
parameter estimate; with a positive gain it is not a separate positive
efficiency threshold. Parameter-free prototypes are handled separately.

Among eligible candidates for a pair, the implementation prioritizes
validation gain, then its fixed encoding preference and estimated
efficiency. At test time, an admitted specialist is used only if the base
model's top two labels match its pair and $m(x)$ is at or below its selected
threshold. Otherwise EmotiQ abstains from intervention and keeps the base
prediction.

\section{Results}
\subsection{Classical Baselines}
We compare 26 non-quantum configurations: ten word TF-IDF models, two
character-aware TF-IDF models, ten models on 96-dimensional SVD features,
and four heads on frozen DistilBERT embeddings. The classifiers include
linear models, support vector machines, Naive Bayes, nearest-neighbor
methods, multilayer perceptrons, and tree ensembles.
Table~\ref{tab:baselines} reports the strongest model in each family.
DistilBERT with logistic regression leads at 0.608 mean macro-F1.
Word-plus-character TF-IDF with a linear SVM is close at 0.583, while the
best SVD-96 model reaches 0.424.

\begin{table}[t]
\centering
\caption{Strongest classical model in each family. F1 is averaged
across the six datasets.}
\label{tab:baselines}
\footnotesize
\setlength{\tabcolsep}{3pt}
\begin{tabular}{@{}lrr@{}}
\toprule
Best model in family & Configs. & Mean F1 \\
\midrule
Frozen DistilBERT + logistic regression & 4 & \textbf{0.608} \\
Word + character TF-IDF + linear SVM & 1 & 0.583 \\
Character TF-IDF + linear SVM & 1 & 0.582 \\
Word TF-IDF + Ridge & 10 & 0.544 \\
SVD-96 + RBF SVM & 10 & 0.424 \\
\bottomrule
\end{tabular}
\end{table}

\subsection{Quantum Feature Screening}
We first test whether compact quantum-inspired representations contain
useful complementary signal. The angle-trigonometric map concatenates the
four inputs with their sine and cosine values, giving 12 features. The
$ZZ$ map adds two nonlinear terms for each of the six input pairs, giving
24 features. Logistic regression uses these explicit maps; the $ZZ$ kernel
SVM instead uses their inner product kernel. A fidelity-prototype candidate
normalizes squared similarities to class prototypes into probabilities.
These screening maps are implemented classically and should be
distinguished from the trainable quantum circuits used later.

Candidate probabilities are blended with the base on margin regions selected by validation, after restricting the candidate distribution to the base
model's top two or three labels. Table~\ref{tab:screening} shows that $ZZ$
features provide the largest mean gain, +0.0030, and improve four datasets.
The $ZZ$ kernel follows at +0.0025, and angle features at +0.0013.
Fidelity prototypes decrease mean macro-F1 by 0.0063 despite improving two
datasets. The largest per-dataset screening gains occur with the $ZZ$
kernel on AI-Roadmap (+0.0227) and Super Emotion (+0.0176). The strongest
screened results on Bulpara-8 and Emotion20 remain slightly below the
base.

\begin{table}[t]
\centering
\caption{Comparison of quantum-inspired feature representations.}
\label{tab:screening}
\footnotesize
\setlength{\tabcolsep}{5pt}
\begin{tabular}{@{}lrrr@{}}
\toprule
Encoding & Mean F1 & Mean $\Delta$F1 & Improved \\
\midrule
$ZZ$ features & \textbf{0.582} & \textbf{+0.0030} & \textbf{4/6} \\
$ZZ$ kernel SVM & 0.582 & +0.0025 & 3/6 \\
Angle-trigonometric & 0.580 & +0.0013 & 3/6 \\
Fidelity prototypes & 0.573 & $-0.0063$ & 2/6 \\
\bottomrule
\end{tabular}
\end{table}

\subsection{Top-k Ambiguity Heads}
Ambiguity heads refine a probability distribution. We compare logistic regression on
four PCA features, a classical random Fourier feature (RFF) control, and a
head that adds re-uploading quantum features. RFF approximates a classical
radial basis function kernel. Validation selects
the margin threshold, a blend weight, and $k\in\{2,3\}$. The selected
candidate probabilities are restricted to the base model's top-$k$ labels
and blended with its probabilities only within that margin region.

As Table~\ref{tab:ambiguity} shows, the classical head gives the largest
mean residual gain (+0.0055), followed by RFF (+0.0017) and the quantum
head (+0.0015). All three lose substantial performance when used alone:
their mean changes range from $-0.236$ to $-0.248$. This contrast is central
to EmotiQ's design. Compact heads can improve selected uncertain decisions
even though they discard too much information to replace the full base.

\begin{table}[t]
\centering
\caption{Compact ambiguity heads as gated corrections or standalone
classifiers. All gains are relative to the matched DistilBERT base.}
\label{tab:ambiguity}
\footnotesize
\setlength{\tabcolsep}{3pt}
\begin{tabular}{@{}llrrr@{}}
\toprule
Head & Policy & Mean F1 & Mean $\Delta$F1 & Improved \\
\midrule
Classical & Gated top-$k$ & \textbf{0.585} & \textbf{+0.0055} & 5/6 \\
RFF & Gated top-$k$ & 0.581 & +0.0017 & 3/6 \\
Quantum & Gated top-$k$ & 0.581 & +0.0015 & 4/6 \\
\midrule
Classical & Standalone & 0.331 & $-0.248$ & 0/6 \\
RFF & Standalone & 0.343 & $-0.236$ & 0/6 \\
Quantum & Standalone & 0.341 & $-0.238$ & 0/6 \\
\bottomrule
\end{tabular}
\end{table}

\subsection{Performance of the EmotiQ Framework}
In a supporting experiment, we train specialists to distinguish emotion pairs that the base model often confuses on validation data. We use validation performance to choose between angle-encoded circuits with six or eight qubits. These pairwise quantum specialists gain +0.0036 mean macro-F1. Their largest
improvements are on BRIGHTER (+0.0099) and AI-Roadmap (+0.0076), while
Emotion20 declines by 0.0018. Classical and RFF specialists remain
competitive. This motivates adding explicit reliability checks to the
boundary-specific design.

Table~\ref{tab:progression} places these results alongside the final
framework. Full EmotiQ uses the
four-qubit specialist bank with admission and abstention. It improves mean
macro-F1 by +0.0076, is positive on five datasets, and is nonnegative on all
six in this evaluation.

\begin{table}[t]
\centering
\caption{Summary of residual mechanisms under their respective
protocols.}
\label{tab:progression}
\footnotesize
\setlength{\tabcolsep}{3pt}
\begin{tabular}{@{}lrrr@{}}
\toprule
Mechanism & Mean $\Delta$F1 & Improved & Nonneg. \\
\midrule
Best encoding ($ZZ$ features) & +0.0030 & 4/6 & 4/6 \\
Best ambiguity head (classical) & +0.0055 & 5/6 & 5/6 \\
Pair quantum specialist & +0.0036 & 5/6 & 5/6 \\
\textbf{Full admitted EmotiQ} & \textbf{+0.0076} & \textbf{5/6} & \textbf{6/6} \\
\bottomrule
\end{tabular}
\end{table}
\begin{table}[t]
\centering
\caption{EmotiQ results on held-out test data.}
\label{tab:dataset-results}
\small
\setlength{\tabcolsep}{5pt}
\begin{tabular}{@{}lrrr@{}}
\toprule
Dataset & Macro-F1 & $\Delta$F1 & Active pairs \\
\midrule
AI-Roadmap & 0.446 & +0.0058 & 2 \\
BRIGHTER & 0.160 & +0.0217 & 2 \\
Bulpara-8 & 0.513 & 0.0000 & 0 \\
Emotion20 & 0.657 & +0.0034 & 2 \\
Super Emotion & 0.489 & +0.0069 & 2 \\
Tanaos-8 & 0.783 & +0.0077 & 3 \\
\bottomrule
\end{tabular}
\end{table}

Table~\ref{tab:dataset-results} reports the final dataset-level scores.
BRIGHTER has the largest gain (+0.0217), followed by Tanaos-8 (+0.0077),
Super Emotion (+0.0069), AI-Roadmap (+0.0058), and Emotion20 (+0.0034).
On Bulpara-8, no candidate passes admission, so the final predictions
exactly match the base. 

Admission also changes the outcome within the final experiment.
Table~\ref{tab:policy} compares the admitted policy with applying each
family to all examples whose top-two pair matches a selected boundary,
without admission or a restrictive margin threshold. Every unrestricted
family has a negative mean gain, from $-0.0004$ for re-uploading to
$-0.0253$ for prototypes. The admitted policy combines candidates
selectively and has a positive mean gain.

\begin{table}[t]
\centering
\caption{Effect of the final admission policy. Unrestricted variants
intervene whenever the top two pair matches a selected boundary.}
\label{tab:policy}
\footnotesize
\setlength{\tabcolsep}{4pt}
\begin{tabular}{@{}lrrr@{}}
\toprule
Policy / specialist & Mean F1 & Mean $\Delta$F1 & Improved \\
\midrule
\textbf{Admitted / selected} & \textbf{0.508} & \textbf{+0.0076} & \textbf{5/6} \\
Unrestricted / re-uploading & 0.500 & $-0.0004$ & 1/6 \\
Unrestricted / angle & 0.500 & $-0.0005$ & 4/6 \\
Unrestricted / shared re-uploading & 0.497 & $-0.0031$ & 2/6 \\
Unrestricted / kernel & 0.479 & $-0.0213$ & 1/6 \\
Unrestricted / prototype & 0.475 & $-0.0253$ & 1/6 \\
\bottomrule
\end{tabular}
\end{table}

\subsection{What the Admission Gate Rejects}
Table~\ref{tab:admission} gives two accepted and two rejected candidates.
The AI-Roadmap angle specialist improves validation macro-F1 with precision
0.75 and zero observed harm. Shared re-uploading on Tanaos-8 also passes.
In contrast, a BRIGHTER re-uploading candidate gains +0.0309 but has a harm
rate of 0.33, so it is rejected. The Bulpara-8 kernel candidate is rejected
for the same reason. A positive net gain can hide damaging changes to
previously correct decisions. The harm check makes that tradeoff explicit.

\begin{table*}[t]
\centering
\caption{Representative validation admission decisions. Precision and harm are measured over changed predictions.}
\label{tab:admission}
\small
\setlength{\tabcolsep}{6pt}
\begin{tabular}{@{}lllrrrl@{}}
\toprule
Dataset & Emotion boundary & Specialist & $\Delta$F1 & Precision & Harm & Decision \\
\midrule
AI-Roadmap & anger vs.\ disgust & Angle & +0.0107 & 0.75 & 0.00 & Accept \\
Tanaos-8 & joy vs.\ excitement & Shared re-uploading & +0.0130 & 1.00 & 0.00 & Accept \\
BRIGHTER & joy vs.\ sadness & Re-uploading & +0.0309 & 0.67 & 0.33 & Reject: harm \\
Bulpara-8 & joy vs.\ trust & Polynomial kernel & +0.0134 & 0.80 & 0.20 & Reject: harm \\
\bottomrule
\end{tabular}
\end{table*}

\section{Discussion and Limitations}
The results support a practical role for compact quantum specialists in improving difficult emotion decisions. EmotiQ combines quantum and classical candidates through validation-guided selection, allowing quantum components to contribute where they demonstrate useful, reliable gains. This hybrid design uses specialists for targeted corrections while retaining the strong classical base prediction whenever no specialist qualifies.

Our evaluation provides a controlled comparison across six datasets using single-label examples and fixed sampled splits. The observed improvements on five datasets, with the sixth preserved through abstention, provide a foundation for broader testing. The current study uses the same validation split for boundary discovery and threshold selection, with BRIGHTER providing 92 validation examples. Future work can strengthen the evaluation through larger validation sets, full dataset experiments, and repeated runs across random seeds with confidence intervals. Testing on new domains will further assess how well the selection policy generalizes.

The four qubit design keeps the simulated specialists compact and provides a practical starting point for hybrid quantum classical emotion classification. The current parameter estimates focus on the specialist modules. Future hardware studies can evaluate latency, noise sensitivity, and energy use. Together with closely matched classical comparisons, these studies can clarify where quantum specialists provide the most useful improvements relative to their computational cost.

\section{Conclusion}
This paper investigates whether quantum models can contribute useful information to text emotion classification when deployed as residual specialists. We propose EmotiQ, a framework that identifies difficult emotion boundaries from validation behavior and selectively applies boundary-specific specialists only when they satisfy predefined reliability and efficiency criteria. Experiments across six emotion classification datasets showed that individual quantum mechanisms can provide modest improvements, but their effectiveness varies substantially across datasets and emotion boundaries. Among the evaluated approaches, interaction-based quantum encodings such as $ZZ$ feature maps and kernels were the most consistently effective. The full EmotiQ framework achieved larger and more reliable gains than any individual residual mechanism, improving mean macro-F1 by +0.0076 over a frozen DistilBERT baseline while remaining non-negative across all datasets through selective abstention. The results suggest that a promising direction for current quantum NLP methods is to apply them as specialized components that address difficult low-confidence decisions. 

\bibliographystyle{IEEEtran}
\bibliography{references}
\end{document}